\documentclass[sensors,article,accept,pdftex,moreauthors]{Definitions/mdpi}

\firstpage{1}
\pubvolume{23}
\issuenum{2}
\articlenumber{614}
\pubyear{2023}
\copyrightyear{2023}
\externaleditor{Academic Editor: Marco Picone}
\datereceived{28 October 2022}
\daterevised{13 December 2022}
\dateaccepted{30 December 2022}
\datepublished{5 January 2023}
\hreflink{https://doi.org/10.3390/s23020614} 
\pdfoutput=1

\usepackage[T1]{fontenc}
\usepackage{graphicx}
\usepackage{esvect}
\usepackage{enumitem}
\usepackage{amssymb}
\usepackage{amsmath}
\usepackage{amsthm} 

\usepackage{algpseudocode}
\usepackage{algorithm}
\usepackage{eurosym}
\usepackage{wasysym}

\algnewcommand\algorithmicforeach{\textbf{for each}}
\algdef{S}[FOR]{ForEach}[1]{\algorithmicforeach\ #1\ \algorithmicdo}

\usepackage{dsfont}
\newcommand{\Set}[1]{{#1}}
\newcommand{\Sum}[0]{\sum}
\newcommand{\Expec}[1]{\mathds{E}[#1]}

\Title{Evaluating Collaborative and Autonomous Agents in Data-Stream-Supported
  Coordination of Mobile Crowdsourcing}
\TitleCitation{Evaluating Collaborative and Autonomous Agents in
  Data-Stream-Supported Coordination of Mobile Crowdsourcing}

\Author{{Ralf Bruns}
  $^{1,\dagger}$\orcidA{}, Jeremias {D\"otterl} $^{1,\dagger}$, {J\"urgen}
  Dunkel $^{1,}${*}
  $^{,\dagger}$\orcidB{} and Sascha Ossowski $^{2,\dagger}$\orcidC{}}

\AuthorNames{Ralf Bruns,  Jeremias {D\"otterl}, {J\"urgen} Dunkel and Sascha
  Ossowski}
\AuthorCitation{Bruns, R.;  {D\"otterl}, J.; Dunkel, J.; Ossowski, S.}

\address{%
$^{1}$ \quad Computer Science Department, Hannover {University}
of Applied Sciences and Arts, \linebreak 30459 Hannover, Germany;
{ralf.bruns@hs-hannover.de (R.B.); doetterlj@gmail.com (J.D.)}
\\
$^{2}$ \quad {CETINIA}
, University Rey Juan Carlos, {Móstoles, 28933 Madrid,}
Spain; sascha.ossowski@urjc.es}

\corres{{Correspondence}
  : juergen.dunkel@hs-hannover.de}

\firstnote{These authors contributed equally to this work.}

\abstract{Mobile crowdsourcing refers to systems where the completion of tasks
  necessarily requires physical movement of crowdworkers in an on-demand
  workforce. Evidence  suggests that in such systems, tasks often get  assigned
  to crowdworkers who struggle to complete those tasks successfully, resulting in
  high failure rates and low service quality.
  A promising solution to ensure higher quality of service is to continuously
  adapt the assignment and respond to failure-causing events by transferring
  tasks to better-suited workers who use different routes or vehicles.
  However, implementing task transfers in mobile crowdsourcing is difficult
  because workers are autonomous and may reject transfer requests. Moreover, task
  outcomes are uncertain and need to be predicted.
  In this paper, we propose different mechanisms to achieve outcome prediction
  and task coordination in mobile crowdsourcing.
  First, we analyze different data stream learning approaches for the prediction
  of task outcomes. Second, based on the suggested prediction model, we propose
  and evaluate two different approaches for task coordination  with different
  degrees of autonomy: an opportunistic approach for crowdshipping with
  collaborative, but non-autonomous workers, and a market-based model with
  autonomous workers for crowdsensing.}


\begin{document}

\section{Introduction}
\label{sec:introduction}

Explicit  crowdsourcing  is a problem-solving paradigm in which  tasks are
outsourced to an on-demand workforce of private individuals. Mobile
crowdsourcing is often considered as a specific case of explicit crowdsourcing,
where  the outsourced tasks are spatial in character, and interested
individuals use their personal mobile devices to temporarily join the crowd,
find nearby tasks, complete the tasks, and earn rewards \citep{alt2010,
	doetterl2021, yan2009}.
\linebreak Mobile crowdsourcing has been used for a wide range of purposes:
\begin{itemize}
	\item \emph{{Crowdsensing}
	      } systems \citep{ganti2011, ma2014} recruit crowdworkers to
	      gather sensor data at locations of interest. There are many applications of
	      crowdsensing in the social sector, the environmental sector, and the
	      infrastructure sector, among others. Examples include measurements of air
	      quality~\cite{dutta2009} and noise pollution~\citep{qin2016, zappatore2016},
	      monitoring of road \linebreak traffic  \cite{hull2006} and road surface
	      conditions  \citep{eriksson2008, mohan2008, sattar2018}, as well as photo
	      acquisition \cite{osawa2017}. \linebreak In these scenarios, data are obtained
	      from many different locations and then used to analyze, maintain, and improve
	      the monitored infrastructure or environment.

	\item \emph{{Crowdshipping}} systems~\cite{sadilek2013} tackle the
	      problem of last-mile parcel delivery by a crowdsourcing approach, i.e., to
	      service the delivery leg from the logistics company's parcel hub to the
	      recipients' home. Interested individuals can join to act as couriers. The aim
	      is to use spare capacities in private vehicles and piggyback on trips that are
	      made by private citizens anyway.
	      In exchange for monetary rewards, the interested individuals pick
	      up parcels (or any other kind of freight) and bring them to their destinations,
	      altering their original route only for minor detours~\cite{giret2018}.
\end{itemize}

The crowd has many advantages compared to permanent staff or fleets of
professional drivers. The crowdworkers are spatially distributed and can solve
many tasks by taking short detours from their original routes, which is fast,
cost-efficient, and sustainable. \linebreak Furthermore, the crowd is a
scalable workforce that can be contracted on a task-by-task basis, which makes
the crowd suitable for short-term projects that would be impractical or costly
otherwise.

Unfortunately, under some conditions, mobile crowdsourcing fails to meet all
expectations. Tasks are frequently assigned to crowdworkers who later face
difficulties to complete these tasks successfully, resulting in high failure
rates, unreliable system behavior, and low service quality. Task execution
takes place in highly dynamic urban environments with many sources of
disruption such as traffic and vehicle issues. From traditional delivery
fleets, we know that unexpected events related to traffic, vehicles, and
weather can result in large delays, high costs, and inferior customer service
\cite{zeimpekis2005}. For instance, traffic jams can introduce huge delays to
travel plans of workers moving by car. Weather conditions may put delivery
deadlines at risk, especially if the means of transportation is a bicycle.
\linebreak Furthermore, during task execution, crowdworkers can be affected by
unexpected situations, distraction, inexperience, tiredness, or competing
obligations. For instance, fatigue can be a reason for a worker walking or
cycling to deliver parcels later than \linebreak expected \cite{doetterl2021}.
Conventional crowdsourcing provides evidence indicating that tasks are
frequently abandoned \citep{han2019a, han2021}. Such events can result in the
delayed completion or failure of tasks. To ensure high service quality and
long-term worker engagement, task failures must be prevented.

A possible method to prevent task failures, which has been successfully
exploited in related domains \citep{cortes2010, billhardt2014event}, is to
respond to failure-provoking events by strategically transferring tasks to more
promising workers. If one worker is unwilling or unable to continue a task,
another worker nearby may be willing and capable to complete the task instead.
Task transfers are a promising technique because the crowd is heterogeneous.
\linebreak Crowdworkers use different transportation modes; they have different
skills, experience levels, and preferences, and they have different
destinations, routes, and travel patterns. Exploiting this heterogeneity, tasks
that are executed too slowly by pedestrians can be executed more quickly by car
drivers. On the other hand, bikes are less affected by road congestion and can
travel under such conditions faster \cite{maes2012}. Hence, car drivers stuck
in traffic may transfer their tasks to unaffected cyclists. Task transfers
allow tasks to be executed faster and the system to behave more reliably.
However, task transfers among crowdworkers have to deal with two main
challenges:
\begin{itemize}
	\item \emph{{Outcome uncertainty:}} While a task is being performed by
	      a crowdworker, the outcome of that task is uncertain, and it is generally
	      unknown whether the task could be successful. However, the decision to transfer
	      a task	depends largely on the likelihood that the task can be completed
	      successfully.  Therefore, to enable task transfers, task outcomes must be
	      accurately predicted.

	\item \emph{{Coordination:}} Task transfers between workers must be
	      coordinated. In general, the crowdworkers involved in a task transfer may have
	      different levels of autonomy. \linebreak Here, we distinguish between
	      \emph{{collaborative workers}}, who must perform a transfer proposed by the
	      system in any case and \emph{{autonomous workers}}, who can decide for
	      themselves which task transfers will be carried out. Only if the task transfers
	      are in line with the individual interests of the autonomous workers will they
	      find acceptance and be executed by the workers. Even if transfers are efficient
	      from the system's point of view, autonomous workers will reject them if they
	      are not beneficial to themselves. \end{itemize}

In this paper we argue that supporting peer-to-peer task transfers among
crowdworkers is a promising approach to prevent task failures in mobile
crowdsourcing. In the following, we propose different mechanisms to achieve
outcome prediction and task coordination. In particular, we set out from
outcome prediction mechanisms  outlined \linebreak in \cite{bruns2022} and
compare a collaborative task coordination model to a market-based mechanism
geared towards an open system of autonomous agents in line with
\cite{doetterl2021}.

More specifically, our main contributions are:
\begin{itemize}
	\item To mitigate the outcome uncertainty, we propose a machine
	      learning approach on data streams that can predict task outcomes.
	\item Based on the suggested prediction model, two different approaches
	      to task coordination are proposed:
	      \begin{enumerate}[label=(\roman*)]
		      \item First, we present how non-autonomous collaborative
		            workers cooperate by means of task transfers in the crowdshipping domain.
		      \item We then propose a market-based model for autonomous
		            workers using a crowdsensing application scenario as an example.  In this
		            approach, autonomous crowdworkers can perform or reject tasks according to
		            their personal interests.
	      \end{enumerate}
\end{itemize}

The paper is organized as follows. In Section \ref{sec:related}, we discuss
some related work.  \linebreak Then, Section \ref{sec:crowdshipping} outlines
how uncertainty can be tackled using continuously monitored sensor data and
data stream learning. It presents a crowdshipping approach based on
collaborative crowdworkers.
In Section \ref{sec:crowdsensing}, we present how worker autonomy can be
implemented in a crowdsensing scenario by coordinating task transfers through a
market-based negotiation mechanism.  Finally, we conclude the paper pointing to
future lines of work in Section \ref{sec:conclusion}.

\section{Related Work}
\label{sec:related}

So far, task transfers have been insufficiently addressed in the context of
mobile crowdsourcing. Only for the crowdshipping domain do a few publications
exist with the explicit aim to introduce task transfers. One of the earliest
proposals in this direction was made by \citet{sadilek2013}. In their  system,
parcels can be transferred between workers if there is {an overlap in space and
		time}. They view mobile crowdsourcing with task transfers primarily as a
routing and planning problem that can be solved with graph planning algorithms.

The assumption of obedient workers is a recurring theme throughout the
crowdshipping
literature~\citep{sadilek2013,chen2018multihop,raviv2018,sampaio2018,sampaio2020,voigt2021}.
For example, \citet{chen2018multihop} study an offline setting where all
delivery requests and shippers are known upfront.
Through offline planning, a matching between parcels and drivers is computed.
However, such planning algorithms assume a fleet of employees that can be
controlled and are hence not directly applicable to crowds of autonomous,
self-interested workers.

These publications---if at all---assume only a limited notion of worker
autonomy. Often, workers must declare their temporal availability upfront and
are assumed to accept any task that lies within the declared availability
\citep{sampaio2018,sampaio2020}.
Sometimes, simple constraints are added, such as vehicle capacity
\citep{chen2018multihop,raviv2018}, a maximal detour length
\citep{chen2018multihop,voigt2021}, or a maximal number of transfers
\citep{voigt2021}.

Up to now, few multi-agent proposals have existed that aim to incorporate task
transfers into crowdshipping.
\begin{itemize}
	\item In \citet{rodriguez2018}, transfers are performed when two
	      drivers benefit financially, and redundant trips can be avoided.
	      The re-assignments are proposed by a plan manager, which has
	      knowledge about all the delivery tasks, drivers, and their properties.

	\item In \citet{giret2018} and \citet{rebollo2018}, multi-agent systems
	      are combined with complex network-based algorithms to route parcels through the
	      crowd.
	      The proposed system maintains live information about the city's
	      transportation network and the couriers' GPS locations.

	\item In \citet{doetterl2020b}, we presented an agent-based approach
	      for delivery delay prevention in crowdshipping. When a delay is predicted, a
	      parcel transfer is proposed that both the current deliverer and the prospective
	      new deliverer are allowed to reject.
\end{itemize}

Besides the challenge of worker autonomy in accepting or rejecting a transfer
request, a second issue remains insufficiently addressed: many crowdshipping
transfer mechanisms operate on data that is assumed to be available upfront and
to remain valid during the entire plan execution.
Various proposals perform offline planning, i.e. a plan is precomputed based on
data that is assumed to be provided to the algorithm upfront
\citep{chen2018multihop,sampaio2018,sampaio2020,voigt2021}. \linebreak
However, in practice unexpected events can interfere with the plan execution.
To react to unexpected events and plan deviations, at least the incorporation
of live data is necessary, as proposed by \citet{giret2018} and
\citet{rebollo2018}. While transfer methods based on live data already have
great advantages over offline methods, they can at most be reactive. Only when
the event has already occurred can a reaction be initiated.

To prevent task failures proactively, task outcomes must be predicted. While a
task is being executed, the outcome of that task is uncertain---during its
execution, it is generally unknown whether the task is going to succeed.
However, the decision of whether a task should be transferred strongly depends
on the anticipated outcome.
To coordinate effective quality-improving task transfers, task outcomes must be
accurately predicted.

The combination of crowdshipping with outcome prediction has received little
attention so far.
One work in this direction is authored by \citet{habault2018}, who suggest
using machine learning to predict delays in a crowdsourced food delivery
platform. \linebreak However, their proposal strongly differs from our
approach: in their proposal, delay prevention is attempted via rerouting---not
via task transfers.

In \citet{doetterl2020b}, we presented an agent-based approach that predicts
delivery delays from the workers' smartphone sensor data. For the delay
prediction, the agents use data stream processing and data stream mining. Based
on these predictions, the agents negotiate task transfers. This article further
extends this work.

We conclude that the literature on task transfers in mobile crowdsourcing to a
large degree focuses on planning algorithms.
However, offline planning algorithms are unable to react to unexpected events
and plan deviations, which regularly occur during plan execution. Existing
papers that use task transfers tend to neglect the crowdworkers autonomy and
the outcome uncertainty associated with tasks and transfers. In order to
effectively address outcome uncertainty, accurate task outcome predictions are
a prerequisite.

\section{Crowdshipping with Collaborative Workers}
\label{sec:crowdshipping}

\emph{{Crowdshipping}
} ({{sometimes referred to as crowd logistics}}),
as defined in Section~\ref{sec:introduction}, is a particularly relevant
example of mobile crowdsourcing. In crowdshipping,  the delivery of parcels to
recipients is outsourced to a crowd of individual workers.

A major goal of the crowdsourcer (usually a logistics company) with regard to
the last mile problem is to establish bilateral agreements with crowdworkers
regarding some transportation tasks such that an acceptable quality of service
is ensured. Among others, this implies that parcels are delivered as soon as
possible, and certainly by some deadline.  In conventional mobile
crowdsourcing, however, it appears that tasks are often abandoned, as
crowdworkers usually accept tasks based on their current situation and
environmental circumstances. However, this often happens to be a misjudgment,
and when this becomes apparent (as the context changes with time), the workers
prefer to hand over the task \cite{han2021}.

In the following, we present worker-to-worker parcel transfers as a new
approach to improve service quality in crowdshipping.  A key element in this
respect is the accuracy of the \emph{{service quality prediction model}} that
is used to trigger task transfer decisions.  In
Section~\ref{subsec:collab_predict}, different data stream learning methods are
applied and analyzed for predicting the  delivery success. In
Section~\ref{sucsec:collab_model}, we embed this prediction model into a
collaborative agent-based crowdshipping model where task transfer decisions are
taken locally and with limited overhead (see also~\cite{bruns2022}).

\subsection{Consideration of Outcome Uncertainty by Predicting Delivery
	Success}
\label{subsec:collab_predict}

Parcel transfers should only be triggered if some delivery task is at risk of
failure. \linebreak Consequently, a  key issue is  predicting the success of
the  parcel delivery outcome. Here, we consider that a task has failed when a
worker cannot complete it  before the deadline. This is only one of several
possible key performance indicators for the collaborative case. (for instance
one could also take in account the amount of delay). However, in this paper, we
aim at comparing collaborative and autonomous crowds, and, especially in the
latter case, it seems reasonable to assume that penalties are always applied
(on a binary basis) when a task is not successfully accomplished in time. {We}
assume that in both scenarios,	all assigned tasks are to be completed
necessarily; i.e., delayed tasks cannot just be dropped. This can be assured
through specific normative or social control schemes (very high monetary
penalties, exclusion  from the crowd, loss of reputation, etc.).
Notice that in a dynamic environment, the quality of the prediction with regard
to this performance indicator depends on the situation awareness of the
crowdworkers.

\subsubsection{General Prediction Model}
\label{subsec:general_prediction}

The prediction of delivery success can be considered as a binary classification
problem. For a crowdworker responsible for a certain parcel, it must be decided
whether the parcel can be delivered on time, or in other words, whether the
assignment belongs to the class {DELAYED} or not.
More formally, the hypothesis function $h()$ shown in Equation
(\ref{hypothesis}) must be learned.
\begin{equation}
	\label{hypothesis}
	\hat{y} = h(\vv{x})
\end{equation}
with  the target $\hat{y} \in $ \{DELAYED, NOT\_DELAYED\} and the feature
vector $\vv{x}=(x_1, x_2, \ldots, x_n)$, which contains all available data used
to predict the task delivery outcome.

However, it is not sufficient to  know only whether an assignment between
courier $c$ and parcel $p$ belongs to class $DELAYED$; we also need to quantify
how certain this prediction is, i.e., the exact value of the delay probability
$prob_{delay}(c, p)$.  Fortunately, most machine learning algorithms are able
to provide probability estimates of their predictions. {For} instance, in a
decision tree, if a data item has been traversed to a particular leaf node,
then the predicted delay  probability is the percentage of delayed items out of
all items that have been propagated to that leaf node.
In scikit-learn, decision\_function() and predict\_proba()) provides provides
such prediction probabilities.

The crowdworker can use three types of data to make a realistic prediction
about whether or not a parcel belongs to the {DELAYED} class:
\begin{enumerate}[label=(\roman*)]
	\item \emph{{Worker capabilities}}: each worker has knowledge about her
	      own traveling speed achieved so far, which is given by the current,  mean,
	      maximum, and minimum speed.
	\item \emph{{Parcel delivery state}}: the current delivery state of a
	      parcel is defined by the remaining distance to the destination and the
	      remaining time until the delivery deadline.
	\item  \emph{{Environmental situation}}: the traffic situation relevant
	      for parcel delivery can be characterized by the current traffic situation  on
	      the remaining delivery route. {We} assume that each worker agent can obtain
	      information about the traffic density on its route via a central service.
\end{enumerate}

For evaluating the quality of a machine learning approach, we calculate
well-known  evaluation metrics for classification problems: precision, recall,
and F1-score~\cite{Tharwat2020}:

\begin{itemize}
	\item $Recall = TP/(TP + FN) $ is the percentage of the DELAYED parcels
	      that are classified correctly, {with} true positives (TP),  false positives
	      (FP), and false negatives (FN). An optimal recall of 1.0 means that all DELAYED
	      parcels are correctly classified as DELAYED.
	\item $Precision =  TP/(TP + FP)$  is defined by the percentage of the
	      correct classification. The precision is of value 1.0 when all parcels
	      classified as {DELAYED are DELAYED}.
	\item The \emph{F1-score} is the harmonic mean of the recall and
	      precision values and provides an aggregated measure of classification quality.
	      A perfect F1-score of 1.0 is given when precision and recall both are 1.0.
\end{itemize}

\subsubsection{Learning from Data Streams}
In a crowdshipping system, the prediction of the delivery success must be
learned on a stream of continuously arriving data: whenever two crowdworkers,
one of whom has a parcel, encounter one another, each of them must estimate the
probability with which she could deliver the parcel on time.
Classification is the task of predicting the correct label (here: {DELAYED} or
{{NOT\_DELAYED)} of a parcel $p$ for an unlabeled feature vector $\vv{x_i}$ (as
introduced in Equation~(\ref{hypothesis}).
During the training stage, the classification algorithm observes a data stream
$D$.

\begin{equation}
	D = \{(\vv{x_i}, y^p)|i = 0, 1, 2, 3, \ldots, m\}
\end{equation}
{where} the $i$-th training data item $(\vv{x_i}, y^p)$ is the feature vector
$\vv{x_i}$ with the corresponding true target label $y^p$.
The machine learning algorithm uses this training data to learn a prediction
model. This model can be used to predict the still unknown and most probable
label $\hat{y}^p$ for a newly arriving feature vector $\vv{x}_{new}$.

Data streams exhibit some characteristic
features~\citep{babcock2002,bruns2015}:
\begin{itemize}
	\item  Data arrives continuously online,
	\item  Data streams are potentially infinite and cannot be held
	      completely in memory,
	\item  Data must be processed when they arrive and cannot be retrieved
	      again unless they are permanently stored explicitly,
	\item Data adapt to temporal changes: concept drift.
\end{itemize}

When learning on data streams \cite{gama2010}, data cannot be clearly separated
between training, evaluation, and testing data (as is the case with batch
learning). Instead, when predicting whether a parcel will arrive delayed, the
correct result is not available until a later time.  For each data item, data
stream learning goes through a processing cycle  '\emph{predict -> fit model ->
	evaluate}' with the following  steps:
\begin{enumerate}
	\item Obtain an unlabeled data item $\vv{x_i}$,
	\item For $\vv{x}_i$: make a prediction $\hat{y}^p = h(\vv{x}_i)$ for
	      parcel $p$ using the current model $h()$,
	\item Determine the true label $y^p$ for $\vv{x}_i$ (here: when at a
	      later time it is known whether parcel $p$ is delayed),
	\item Use the new correct pair $(\vv{x}_i, y^p)$ to train the current
	      model $h()$,
	\item Take the pair $(\hat{y}^p, y^p)$ to update statistics for
	      evaluation of the model quality.
\end{enumerate}

{Following} these steps, we apply an interleaved or \emph{{prequential}}
evaluation approach:
Each item is first used to test the model by making a prediction for this
previously unseen item. Then, the model is updated (trained) with this item as
soon as its label is available.

There are several stream learning algorithms  that adapt well-known
batch-learning classification algorithms  to data streams. Among others, we
applied K-nearest neighbors (KNN), random forest, and Hoeffding trees
\citep{Aggarwal15, gama2010, hulten2001}.

\subsubsection{Experimental Results for Predicting Delivery Outcome}
Using the data streams generated by an agent-based crowdshipping simulator (see
Section~\ref{subsec:simulator}), we conducted various experiments for
predicting the delivery success.
Our machine learning experiments were performed with \emph{{River}}
(\url{https://riverml.xyz/} {(accessed on 12 December 2022))}
that integrates the data stream learning libraries \emph{{scikit-multiflow}}
(\url{https://scikit-multiflow.github.io} {(accessed on 12 December 2022))} and
cr\'{e}me  (\url{https://pypi.org/project/creme/} {(accessed on 12 December
	2022)). }
Table~\ref{tab:collab_prediction} shows our results using the data stream
versions of KNN, random forest, and Hoeffding.

\begin{table}[H]
	\caption{Experimental results for the prediction of the delivery
		success.}
	\centering
	\newcolumntype{C}{>{\centering\arraybackslash}X}
	\begin{tabularx}{\textwidth}{lCCC}
		\toprule
		\textbf{Prediction Model}  & \textbf{Precision} &
		\textbf{Recall}            & \textbf{F1-Score}          \\
		\midrule
		Random Forest {($n$} = 20) & 0.957              & 0.974
		                           & 0.966                      \\
		KNN ($K$ = 10)             & 0.912              & 0.904
		                           & 0.908                      \\
		Hoeffding                  & 0.879              & 0.881
		                           & 0.880                      \\
		\bottomrule
	\end{tabularx}
	\label{tab:collab_prediction}
\end{table}

All applied data stream learning methods behaved almost the same and provided
very good prediction results. In particular, ensemble learning with random
forest	based on $n=20$ trees yielded precision and recall values of better
than 95\%.
Furthermore, appropriate prediction models were learned fast.
Figure~\ref{fig:convergence} shows the convergence of the data stream learning
process considering the F1-score for the three machine learning methods. After
the arrival of about 2000 data items, the F1-score for all methods  reached a
value of about 90\%. After that, the F1-score was almost constant, only for
random forest it was slightly increasing.

\begin{figure}[H]
	\includegraphics[width=0.8 \textwidth]{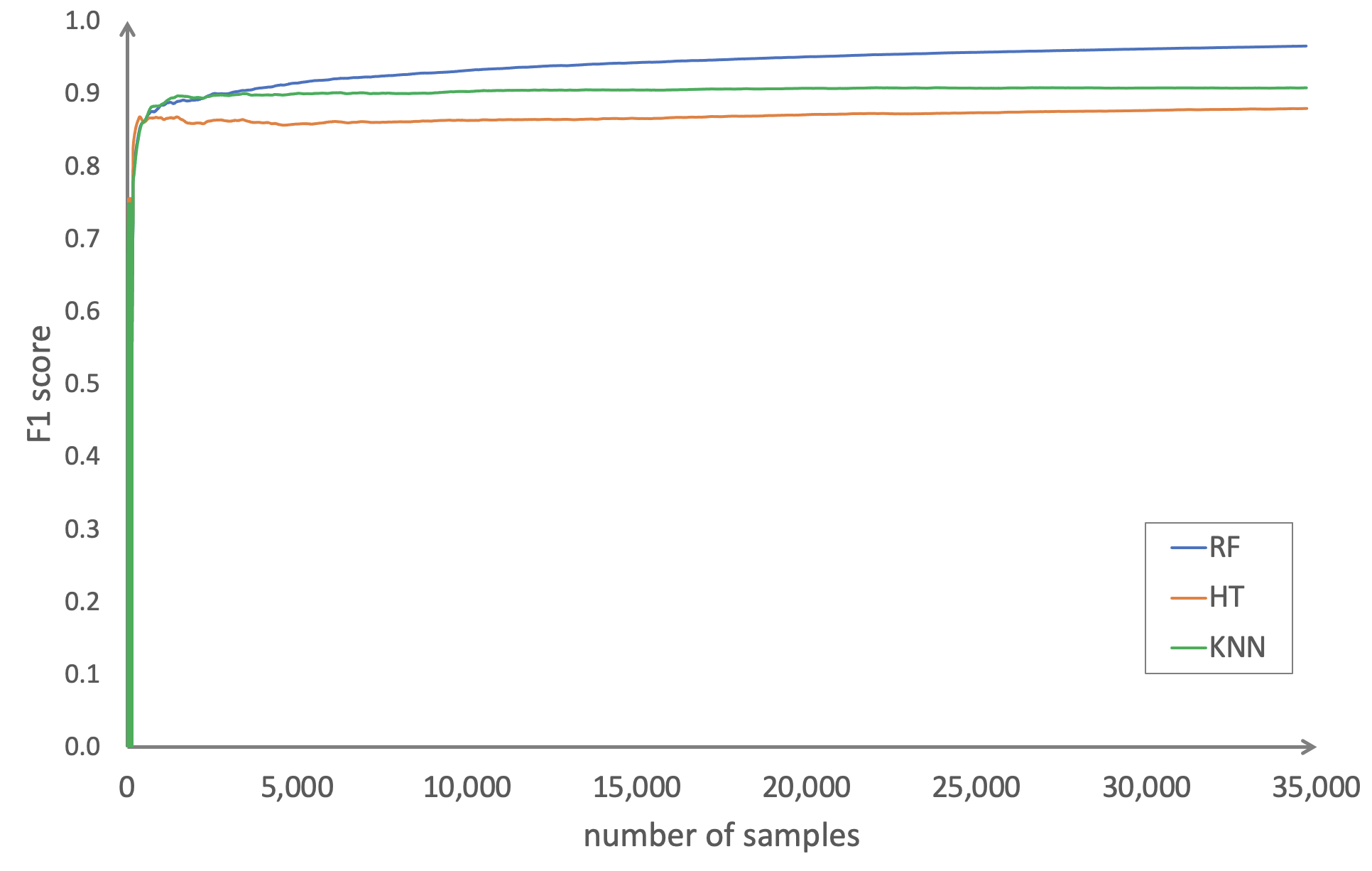}
	\caption{{F1-score} convergence for random forest (RF), Hoeffding (HT),
		and KNN.}
	\label{fig:convergence}
\end{figure}

When deriving a prediction model, we are interested not only in its accuracy
but also in understanding how the predictions are inferred. A feature
importance value can be viewed as a percentage expressing  how much a
particular feature contributes to the prediction.
Feature importance helps us understand which features are mainly used in the
prediction and may allow us to reduce the feature space, making the model
training much faster.

Table~\ref{tab:featureImportance} shows how the importance of the  features
introduced in Section~\ref{subsec:general_prediction} is distributed among the
different types.

\begin{table}[H]
	\caption{Aggregated feature importance.}
	\newcolumntype{C}{>{\centering\arraybackslash}X}
	\begin{tabularx}{\textwidth}{lC}
		\toprule
		\textbf{Feature}
		                & \textbf{Importance} \\
		\midrule
		Worker capability (current, max, min, mean speed)
		                & 0.19                \\
		Parcel delivery state	(remaining time, remaining distance)
		                & 0.54                \\
		Environmental situation (traffic states and distances on
		remaining path) & 0.27                \\
		\bottomrule
	\end{tabularx}
	\label{tab:featureImportance}
\end{table}

Our data stream learning approach is based on many different features. To
provide a better overview, we  aggregated the feature importance values of the
individual features according to three categories by summing up the individual
values. The most important features relate to the delivery state of the parcel
with an overall importance of about (54\%). In particular, the remaining
delivery time is by far the most important feature (36\%).
The features associated with the workers' capabilities, specifically how fast
they are, have an overall importance of 19\%.
The environmental features sum up to about 27\% and describe the remaining path
with the corresponding traffic states.

\subsection{A Collaborative Agent Model for Task Transfers in Crowdshipping}
\label{sucsec:collab_model}

Task transfers are particularly challenging in the crowdshipping domain, as
either crowdworkers  must physically meet at some place at a specific time or
the parcel needs to be stored in a secure place and picked up later by the
newly responsible worker. In traditional fleet management systems such parcel
transfers can be arranged in a top-down manner by the logistics company, but
the decentralized nature of the crowdshipping approach calls for methods that
rely on peer-to-peer interaction \cite{doetterl2020b}.

In the following, we propose an agent-based collaboration model towards the
crowdshipping problem (see also~\cite{bruns2022}). In this model, the option of
performing a transfer of parcel $p$ from its bearer $i$ to another worker $j$
is explored only when the couriers $i$ and $j$ ``encounter'', i.e., when their
locations are close to each other. In this case, the expected quality of
service provided by $i$ performing $p$ is compared to $j$'s service quality
executing $p$, and if the latter exceeds the former, the parcel $p$ is
transferred from $i$ to $j$.

Service quality here does not need to be defined exclusively in terms of
potential task delay but may also consider factors such as fuel cost, CO$_{2}$
footprint, reliability, etc.

The model avoids a major source of complexity: parcel transfers can be
effectively implemented on-the-fly because, as mentioned above, both workers
are already near to one another. Of course, such a stance necessarily implies
less emphasis on strategic issues.

Algorithm~\ref{algorithm} summarizes this collaborative worker agent model for
parcel transfers.  \linebreak It is obvious that it depends on two key
elements:
\begin{enumerate}
	\item A function for predicting the  \emph{service quality} of a worker
	      with respect to a delivery task is needed.
	      The service quality function $F_{qual}(c, p)$ specifies how well
	      a crowdworker $c$ is able to achieve the delivery of a parcel $p$; i.e.,  a
	      high value means better quality.
	\item The concept of \emph{{vicinity}} needs to be specified so as to
	      be able to identify the set $CS_c^t$ of candidate couriers.  Notice that the
	      more restrictive the notion of vicinity is,  the more seldomly couriers
	      encounter one another, and the fewer task transfer opportunities~exist.
\end{enumerate}

For instance, the service quality can be specified by the prediction of the
delivery outcome learned by the prediction model introduced in
Section~\ref{subsec:collab_predict}.
Thus, the service quality function in line 2 can be defined as:
\begin{equation}
	F_{qual}(c, p) = 1 - prob_{delay}(c, p)
\end{equation}

{A} parcel should be transferred to a worker of higher quality, i.e., with a
lower delay probability.

The notion of vicinity is used in Line 5 of the algorithm in order to determine
the set of possible candidate workers. Several different options exist to
specify the vicinity of two workers. A simple and practical model for vicinity
is the definition of a cell-based environment model, where a grid  is placed
over the urban operation area.

Whenever a courier worker with an assigned parcel moves (in discrete time steps
in Lines 4--10), it looks for a more suitable worker in the vicinity.
According to Algorithm~\ref{algorithm}, Line 5, nearby located crowdworker
agents are potential candidates for a parcel transfer. \linebreak By
definition, all agents who are currently in the same grid cell in a time step
as the courier worker are in vicinity to each other and, therefore, are
considered as candidates.
\linebreak The prediction model learned in Section~\ref{subsec:collab_predict}
is now	used to estimate the delivery success probability of  each candidate.
A parcel transfer takes place if at least one of the candidate workers has a
smaller delay probability as the courier worker (Lines 7--8).  \linebreak The
parcel is transferred to that candidate with smallest delay probability. For
the sake of simplicity, it is assumed that candidates do not yet have an
assigned parcel, and transfers do not take any time (duration = 0).

\begin{algorithm}[H]
	\caption{Collaborative parcel transfer algorithm.}
	\label{algorithm}
	\begin{algorithmic}[1]
		\Procedure{Collaborative ParcelTransfer}{}
		\State Define service quality function $F_{qual}(c,p)$;
		\ForEach{time step $t$}
		\ForEach{courier $c$ with parcel $p$}
		\State Determine the candidate set $CS_c^t$,
		\Comment{all couriers $c_i$ in vicinity to $c$}
		\State $cand_{best} = \{c^* \in CS_c^t	\textbar
			F_{qual}(c^*,p)  = \max\limits_{\forall c_i \in CS_c^t}  (F_{qual}(c_i,p))\} $
		\If{$F_{qual}$($c,p$) < $F_{qual}$($cand_{best},p$) }
		\Comment{$candidate~is~better~suited$}
		\State Transfer parcel $p$ from $c$ to $cand_{best}$
		\EndIf
		\EndFor
		\EndFor
		\EndProcedure
	\end{algorithmic}
\end{algorithm}

\subsection{Experimental Evaluation of Collaborative Agent Approach}
\label{subsec:collab_results}
Using the agent-based simulator introduced in Section~\ref{subsec:simulator}
and the prediction models of Section~\ref{subsec:collab_predict}, we conducted
extensive experiments of our proposed collaborative crowdshipping approach.

As the baseline of experiments, we implemented a conventional crowdshipping
approach, where every worker has to deliver its initially assigned parcel
without any possibility of exchange.  In a simulation run comprising 600
parcels,  the conventional approach without any parcel transfers results in
39\% of the parcels being delayed, i.e., 234 of 600 parcels cannot be delivered
in time. The mean completion time over all parcels is 16.28 min.

Table~\ref{tab:transfers} shows the simulation results of our collaborative
crowdshipping model with the different prediction models presented in
Section~\ref{subsec:collab_predict}. It can be seen that the percentage of
delayed parcels can almost be reduced by half (from 39\% to approx. 21\%), and
the mean completion time drops by 25 \% from around 16 min to 12 min.
The fourth column lists the total number of  transfer activities and the fifth
column the number of reassigned parcels. It can been seen that some parcels
have been exchanged more than once. \linebreak For instance,  249 transfers
from a current courier to a candidate worker took place, and 210 parcels were
affected. The last column lists the number of parcels in time or delayed after
being transferred, respectively. For instance, taken the 210 reassigned
parcels, 162 could have been delivered in time after the transfer, whereas 48
parcels were still delayed.

\begin{table}[H]
	\caption{Experimental results of collaborative parcel transfers. }
	\newcolumntype{C}{>{\centering\arraybackslash}X}

	\begin{adjustwidth}{-\extralength}{0cm}
		\begin{minipage}{\fulllength}
			\begin{tabularx}{\textwidth}{lCCCCC}
				\toprule
				\textbf{Prediction Model}  & \textbf{Delay}                   &
				\parbox{2cm}{\textbf{Completion}                                                                            \\ \textbf{Time}}
				                           & \textbf{Transfers}               & \parbox{1.5cm}{\textbf{Reassigned}          \\ \textbf{Tasks} }
				                           & \parbox{1.5cm}{\textbf{In Time/}                                               \\ \textbf{Delayed}  }  \\
				\midrule
				Random Forest {($n$} = 20) & 20\%                             &
				12.55 min                  & 249                              & 210                                & 162/48 \\
				KNN ($K$ = 10)             & 22\%                             &
				12.67 min                  & 256                              & 216                                & 169/47 \\
				Hoeffding                  & 21\%                             &
				11.97 min                  & 227                              & 196                                & 159/37 \\
				Convent. Crowdshipping     & 39\%                             &
				16.28 min                  & --                               & --                                 & --     \\
				\bottomrule
			\end{tabularx}
		\end{minipage}
	\end{adjustwidth}
	\label{tab:transfers}
\end{table}

The three  prediction models applied provide similar results in terms of delay
probabilities and completion times. This was to be expected because the results
of the prediction models in Section~\ref{subsec:collab_predict} also hardly
differ from each other.

Overall, the results demonstrate that  collaborative crowdshipping with
delivery outcome prediction has the potential to significantly improve service
quality of last mile delivery processes.

\subsection{Agent-Based Crowdshipping Simulator}
\label{subsec:simulator}
In order to evaluate the feasibility of our task transfer models, we developed
an agent-based crowdshipping simulator (an extended version of the simulator
presented in~\cite{doetterl2020}). The simulator implements a simplified model
of the environment, in particular how crowdworkers behave and how the traffic
situations change. The simulator can be operated with different task transfer
coordination strategies that realize different levels of agent autonomy.  The
implementation of a concrete strategy can be plugged into the simulator.

To make the simulation as realistic as possible, we employed real-world GPS
data to simulate the physical movements of the agents. We chose  open data of a
bicycle sharing system because of its conceptual similarity to a crowdshipping
system. In both domains, there are users who register in the system, physically
move around the urban area, and log out of the system.
Specifically, we employed the GPS data from the Bike Sharing System of Madrid
(BiciMAD)
(\url{https://opendata.emtmadrid.es/Datos-estaticos/Datos-generales-(1)}
{(accessed on 12 December 2022)).}
The data set includes the rides logged by users and the GPS events recorded
during each ride:  start timestamp,  start location,  end location, and  GPS
traces.  Based on BiciMAD data we simulated parcel delivery in the urban city
center of Madrid. The appearance of a worker agent, its start location and
destination, and its route were derived from the BiciMAD data set.

The simulation runs were conducted with several thousand  worker agents and
several hundred parcels to be delivered. The parcels had randomly selected
origin and destination locations within the operating area.
Figure~\ref{fig:map} shows a screenshot of a simulation run. (Red triangles
visualize remaining, gray triangles already visited tasks. Green circles show
current, gray circles past agent positions).

\begin{figure}[H]
	\includegraphics[width=0.7\textwidth]{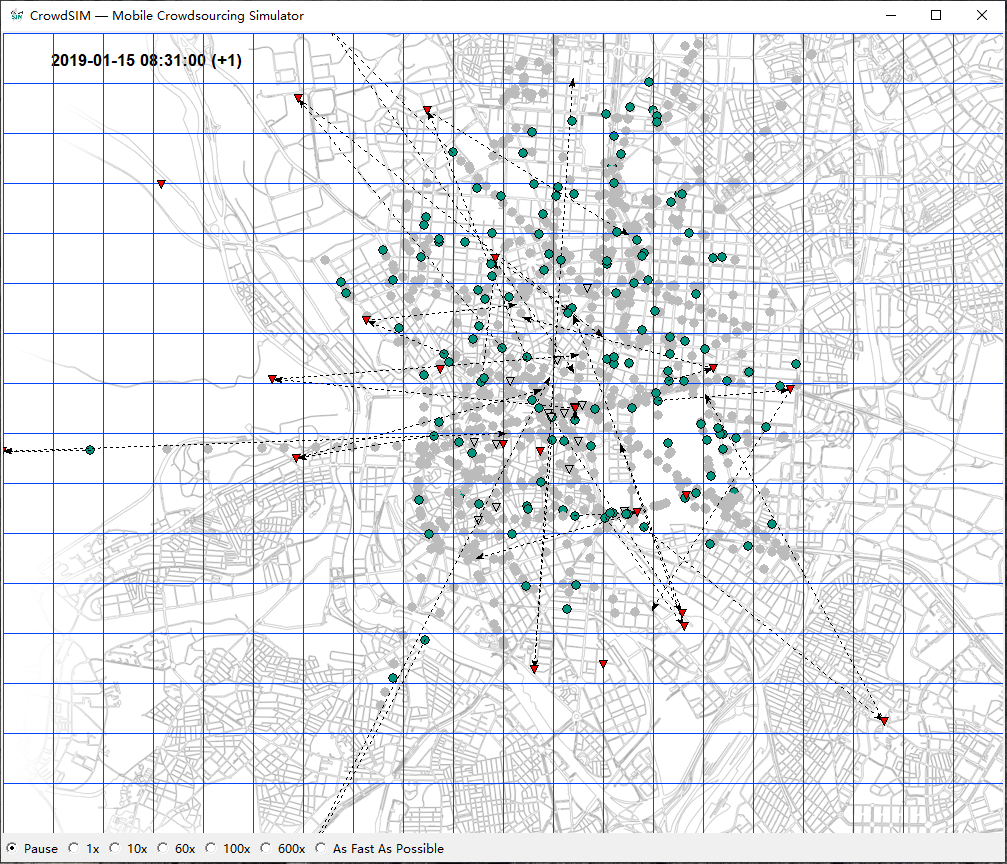}
	\caption{{Agent-based} simulator at work.}
	\label{fig:map}
\end{figure}

\subsubsection{Environmental Model}
Crowdshipping  usually operates in highly populated urban areas. In an urban
area, the speed at which a crowdworker moves is strongly determined by the
current traffic situation.
In order to realize a simple traffic model,  we divided our operating area into
a grid structure.
We assumed that each grid cell can be in one of  the following	traffic states:
\emph{{normal traffic, slow traffic, and traffic jam}}. For crowdworkers
moving on the road, the state of the cells affects their velocity. For
instance, in a cell with state \emph{{traffic jam},} the agents  using the road
do not move at all (travel with 0 km/h). In a cell which is in the state
\emph{{normal traffic}}, all agents travel with their individual preferred
speeds, which is determined by the class to which an agent belongs (see Section
\ref{sec3.4.2}).

Of course, the traffic situation in an urban area changes very dynamically, and
as a consequence, the velocity of road users also  varies over time. We used a
time-discrete Markov chain to model the changing traffic state in a rather
simple way.
In the simulation, the cell states are updated frequently.

\subsubsection{Crowdworker Agents}\label{sec3.4.2}
The crowd consists of individual worker agents who are registered in the system
and interested in delivering parcels.
The model distinguishes three types of crowdworker agents with different
properties: \emph{{walk, bike, and motorbike}}.
Each agent is defined by an arrival time,  start location, end location, and a
route with several predefined  stopover points.  This means, the agents move on
different paths of different lengths.

The walk agents and the bike agents do not move on roads; therefore, their
speed does not depend on the traffic situation. These two agent types move with
a constant speed: walk with 2 m/s  and bike with 5 m/s.
The motorbikes travel on the road, and their speed consequently depends on the
traffic situation of the grid cell where they are  currently located.

For a more detailed description of the simulator and the concrete parameters
used for experimentation, please refer to the crowdshipping
version~\cite{bruns2022}  and crowdsensing version~\cite{doetterl2021}.

\section{Crowdsensing with Autonomous Workers}
\label{sec:crowdsensing}

Crowdsensing systems, as defined in Section~\ref{sec:introduction}, rely on a
crowd of volunteer individuals	collecting sensor data at specific locations.
Crowdsensing inherits many advantages from crowdsourcing, such as access to a
scalable workforce. It is known that crowdsensing has  many advantages over
sensing with fixed sensor installations~\citep{ma2014}.

In this section, we show that task transfers can be used effectively to instill
and maintain coordination in a workforce of \emph{{autonomous workers}}. For
this purpose, we apply the {\emph{Auction-based} \emph{Task} \emph{Transfers}
		(ATT)} approach introduced in~\cite{doetterl2021} to the crowdsensing domain.
Crowdworkers are perceived as mobile agents capable of performing sensing tasks
at their current location (e.g., using their mobile phones). Still, different
from the setting analyzed in Section~\ref{sec:crowdshipping}, here we assume
that crowdworkers are \emph{{individually rational}} and only accept an
additional task when the reward obtained from its successful completion
outweighs the corresponding cost. We show that by allowing such self-interested
crowdworkers to transfer tasks among each other on-the-fly, the overall
performance of the system can be improved without significantly compromising
their autonomy.

It should be noted that as in the crowdsensing domain, all tasks are fully
virtual; they can be easily transferred between workers without the need for
handovers. No parcel or other item has to be handed over between the workers,
so agents do not have to visit a same location. Therefore, as opposed to the
crowdshipping domain discussed in Section~\ref{sec:crowdshipping}, such task
transfers can in principle be performed regardless of the positions of the
involved agents. This implies that once agreed upon by the involved
crowdworkers, task transfers are \emph{{immediate}} and thus cannot fail due to
changing circumstances.

\subsection{Autonomous Crowdsensing with Task Transfers}
In this subsection, we describe our market-based coordination mechanism with
self-interested crowdworkers. Therefore, the concepts of task rewards,
penalties, and expected task outcome are introduced, so as to define the notion
of expected task utility. \linebreak Subsequently, the protocol for task
transfer auctions among crowdsensing agents is described. We finally outline
rational agent strategies for triggering auctions and bidding.

\subsubsection{Expected Utility of Task Sets}
In the crowdsensing domain, each crowdworker~$a_i$ has a utility
function~$U_{i,t}(\Set{S})$, which indicates~$a_i$'s utility for being assigned
a set of sensing tasks~$\Set{S}$. Each sensing task~$\theta_j
	\in \Set{S}$ generates a reward~$r_j$ for the agent completing it
successfully and a penalty~$p_j$ if the crowdworker does not execute the
corresponding service with the required quality. \linebreak Task utilities are
computed as follows:
\begin{equation}
	\begin{split}
		U_{i,t}(\Set{S}) &= -C_{i,t}(\Set{S}) + Rev(\Set{S}) \\
		&= -C_{i,t}(\Set{S}) + \Sum_{\theta_j \in \Set{S}} O_jr_j -
		\Sum_{\theta_j \in \Set{S}} (1-O_j)p_j
	\end{split}
\end{equation}
where~$O_j$ denotes the binary task outcome (successful execution or failure).

This utility function expresses what the crowdworker incurs in costs for
performing the tasks in~$\Set{S}$, which are compensated by the revenue that
the worker obtains by performing them.
The worker's revenue can be negative in cases where the worker fails to
successfully complete some of their tasks and has to pay task penalties. The
revenue~$Rev(\Set{S})$ is computed by summation of all rewards and subtraction
of all penalties. The worker-specific cost function~$C_{i,t}(\Set{S})$ is
subadditive and increases weakly monotonically in the size of S. \linebreak It
models not only the worker's monetary expenses such as gasoline, vehicle
maintenance, and vehicle insurance, etc., but also includes the compensation
that the worker expects for the intangible factors, such as the required effort
and lost time for performing the tasks.

The task set utility~$U_{i,t}(\Set{S})$ can be computed once the outcomes~$O_j$
of all tasks~$\theta_j \in \Set{S}$ are known. For decision making, a
worker~$a_i$ needs access to the expected set task utility, which is given  as:
\begin{equation}
	\begin{split}
		\label{eq:expected-utility-s}
		EU_{i,t}(\Set{S}) &= -\Expec{C_{i,t}(\Set{S})} +
		\Expec{Rev(\Set{S})} \\
		&= -\Expec{C_{i,t}(\Set{S})} + \Sum_{\theta_j \in \Set{S}}
		\Expec{O_j}r_j - \Sum_{\theta_j \in \Set{S}} (1-\Expec{O_j})p_j
	\end{split}
\end{equation}

To determine the expected task set utility in practice, online data stream
learning techniques can be used to compute the expected task
outcomes~$\Expec{O_j}$. For this, we rely on techniques similar to the ones
described in Section~\ref{subsec:collab_predict}.

\subsubsection{Task Transfer Auctions}

Crowdsensing agents are registered to a software platform that sets up a
computational market where sensing tasks can be traded among the agents. If an
agent~$a_i$ believes that it is beneficial for it to give away one of its
sensing tasks~$\theta_j$, it can trigger an online auction. For this, it
announces~$\theta_j$ as the ``good'' to be auctioned to the agents in its
neighborhood~$G_t(\theta_j)$. Figure~\ref{fig:auction-neighborhood} depicts
different neighborhoods of task $\theta_j$ depending on the position~$l_j,_k$
of the agent that it is currently assigned to.

An agent interested in becoming the new assignee of task~$\theta_j$ may then
respond with a bid, indicating the amount~$b$ that it is willing to pay for
this. Notice that the reward~$r_j$ or penalty~$p_j$ are attributed to the agent
that finally completes the task~$\theta_j$, so there is an incentive to ``pay''
for receiving an additional task~$\theta_j$. The auction rules determine the
auction winner~$a_k$, who will become the new assignee of task~$\theta_j$, as
well as the transfer reward~$r_z$ that it has to pay to~$a_i$. In this work, we
use a one-shot second-price (Vickrey) auction for this purpose as it makes
strategic bidding more difficult~\citep{vickrey1961,sandholm1996}.

\begin{figure}[H]
	\includegraphics[width=0.98\textwidth]{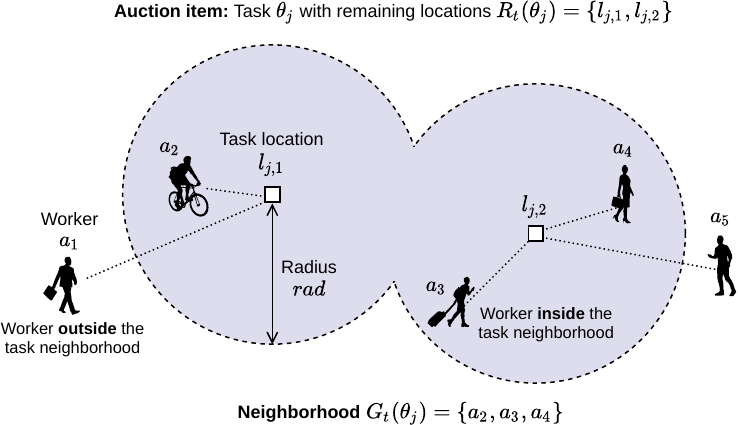}
	\caption[Task Neighborhood Example]{Task neighborhood example.}
	\label{fig:auction-neighborhood}
\end{figure}

\subsubsection{Agent Strategies}

A crowdsensing agent is continuously monitoring its set~$S$ of tasks~$\theta_j$
assigned to it. If, based on the expected task outcome~$\Expec{O_j}$, the
expected utility of giving away~$\theta_j$ plus a potential transfer
reward~$r_z$ exceeds the expected utility of its current set of sensing
tasks~$S$ ($\Delta{eu} > 0$), it will trigger an auction. Still, the transfer
reward~$r_z$ is unknown at the time of launching an auction, and in our
implementation, agents are conservative and only launch an auction
when~$\Delta{eu} > 0$, even with~$r_z=0$.

Similarly, an agent will join an auction as a bidder if, based on the expected
task outcome~$\Expec{O_j}$, the expected utility of adding~$\theta_j$ to its
set of sensing tasks $S$ exceeds the expected utility of $S$ alone. Let $b$
denote the difference between both values, which constitutes its true valuation
of ``winning''~$\theta_j$ in the auction. As we are using Vickrey auctions,
there is no need for the agents to employ a complicated bidding strategy
because, assuming the agents' valuations are independent, the dominant strategy
for all agents is to bid their true valuations. Therefore, we assume our
rational agent's bid value in the online auction just to be $b$.

\subsection{Evaluation}

In this section, we explore how well the ATT approach performs in different
crowdsensing scenarios.
We start by outlining the experimental setup to then present results on delay
percentages as well as average and individual profits.

\subsubsection{Experimental Setup}
For our experiments, we used an instantiation of the simulator described in
Section~\ref{subsec:simulator} and enriched the BiciMAD data set with synthetic
data to simulate the arrival of new tasks over time.
Each task consisted of a set of locations from the crowdsensing operating area.
\linebreak For our experiments, we chose a 1500~m radius in the center of
Madrid, which covered the urban city center where many trips of the BiciMAD
data set (and therefore the simulated agents) pass through. Each task consisted
of three locations, which were aligned to form a chain such that each successor
location was 500~m apart from its predecessor.
All tasks have deadlines, rewards, and penalties: deadlines were set to 30~min
after task occurrence. In our experiments, we assumed that task failures are
undesirable but acceptable with a certain ratio. As a scenario that seems
reasonable, we set the reward and the penalty to the same amount (EUR 5). In
\cite{doetterl2021}, the effect of different reward--penalty combinations was
investigated. As the crowdsensing operating area was rather small, we set the
task neighbourhood radius to be infinite, i.e. task transfer auctions were
broadcast to all agents.

We compare our approach \emph{ATT} with different baselines:
\begin{itemize}
	\item \emph{{Not}} (No transfers):
	      Using this strategy, the crowdworkers do not transfer tasks
	      between each other.
	      Even when tasks are at risk of failing, crowdworkers cannot offer
	      the task to other workers but have to try to finish the tasks themselves.
	\item \emph{{Random}} (Random transfers):
	      Using this strategy, for each time step and task, there is a
	      random chance that the task is transferred to the nearest worker without any
	      tasks.
	      This baseline models a ``blind'' transfer behavior: the transfers
	      are initiated without predictions and the task recipient is chosen without
	      predictions.
	\item \emph{{Forced}} (Forced transfers):
	      Using this strategy, the system identifies the most promising
	      task transfers and executes them without the workers' consent.
	      The system makes predictions about the task outcome, and if a
	      better worker can be identified, for which the task success has a higher
	      probability, the task is transferred to that worker.
	      This baseline is useful to evaluate how the system would perform
	      if the system objective could be maximized if workers were not autonomous.
\end{itemize}

To gain a deeper insight into how \emph{ATT} and the baselines behave in
different settings, we compared them in three scenarios:
\begin{itemize}
	\item \emph{{Scenario 1}} (Base scenario):
	      In this scenario, we generated 50~tasks per hour (600~total) and
	      set the incident probability to~5\%.
	\item \emph{{Scenario 2}} (Increased system load):
	      In this scenario, we generated 100~tasks per hour (1200~total)
	      and set the incident probability to~5\%.
	      The purpose of this scenario was to explore how the different
	      strategies respond to an increase in tasks.
	\item \emph{{Scenario 3}} (Increased environment hostility):
	      In this scenario, we generated 50~tasks per hour (600~total) and
	      set the incident probability to~10\%.
	      The purpose of this scenario was to explore how the different
	      strategies respond to higher disruption levels.
\end{itemize}

For each scenario--strategy pair, we executed 30~simulation runs and report the
mean of delay percentage and worker profits.

\subsubsection{Delay Percentage}

The bar chart in Figure~\ref{fig:barchart-baselines-delays} shows the delay
percentages for the four strategies in the three scenarios.

\begin{figure}[H]
	\includegraphics[width=1.0\textwidth]{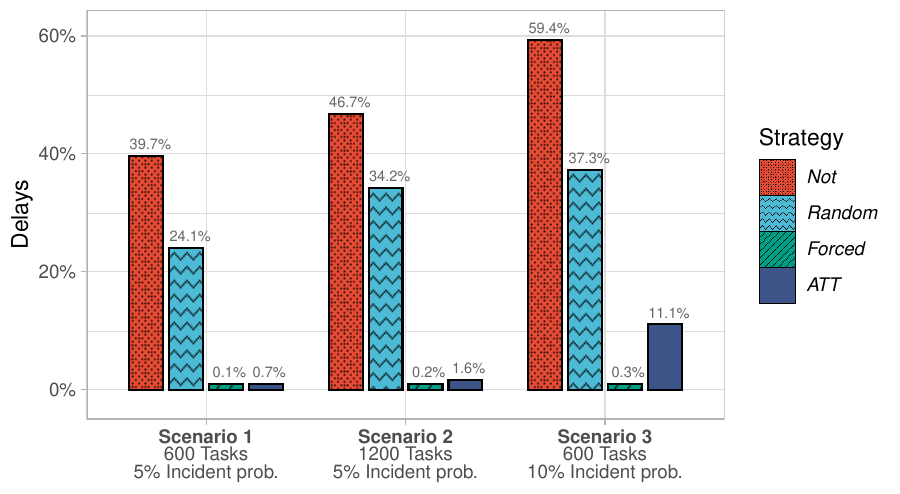}
	\caption[Delays Under Four Strategies in Three Crowdsensing
		Scenarios]{Delays under four strategies in three crowdsensing scenarios.}
	\label{fig:barchart-baselines-delays}
\end{figure}

Across the three scenarios, \emph{{Forced}} and \emph{ATT} performed
significantly better than \emph{Not} and \emph{Random}.
Both \emph{Not} and \emph{Random} struggled in all scenarios and were affected
by higher task counts and incident probabilities.
\emph{ATT} could handle the increase in tasks but started to struggle when the
incident probability was high.
\emph{Forced} produced hardly any delays in any of the three scenarios.

The poor performance of \emph{Not} and \emph{Random} can be explained by the
high incident probability.
The excellent performance of \emph{Forced} and \emph{ATT} shows that task
transfers help to effectively reduce delays.
The superiority of \emph{Forced} and \emph{ATT} compared to \emph{Random} shows
that those transfers must be guided by accurate task outcome predictions;
transfers are only effective if they are based on predictions rather than
randomness.

\emph{Forced} produced fewer delays than \emph{ATT}, which is unsurprising.
While the effectivity of \emph{ATT} was restricted by the workers' autonomy,
\emph{Forced} can initiate arbitrary task transfers that it considers useful
for delay prevention.
In other words, the reason for the superior performance of the \emph{Forced}
strategy is its larger action space: \emph{Forced} can transfer tasks to any
worker, while \emph{ATT} can transfer tasks only to willing workers.

\subsubsection{Average Profits}

We now investigate how \emph{ATT} compares with the baselines in terms of
worker profits.
The mean profits of the participants using the four strategies in the three
different scenarios are shown in Figure~\ref{fig:barchart-baselines-profits}.
Across the three scenarios, \emph{Not} performed worst and yielded high
negative profits.
The \emph{Random} strategy also performed too poorly to achieve positive worker
profits.
On average, both \emph{Forced} and \emph{ATT} provided positive profits to
workers with \emph{ATT} clearly dominating \emph{Forced}.

The negative profits generated by \emph{Not} and \emph{Random} are explainable
by the high number of delays that are observed under these strategies.
The delays result in penalty payments, which cause the negative profits.
\emph{Forced} and \emph{ATT} resulted in very few delays and hence lead to many
reward payments and fewer penalties.
The plot reveals that the negative profits take more extreme values than the
positive profits.
This is because the rewards and penalties paid to and charged from the workers
are further reduced by the costs.
For instance, under \emph{Not} the workers not only suffer penalties but incur
additional costs as well.

\begin{figure}[H]
	\includegraphics[width=1.0\textwidth]{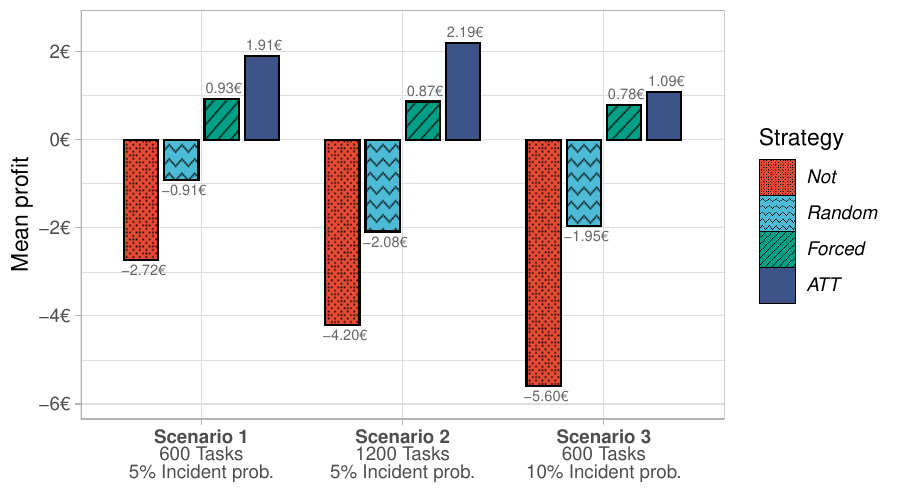}
	\caption[Profits Under Four Strategies in Three Crowdsourcing
		Scenarios]{{The}
		observed profits under four strategies in three crowdsensing
		scenarios.}
	\label{fig:barchart-baselines-profits}
\end{figure}

\emph{ATT} leads to higher profits than \emph{Forced}:
While \emph{Forced} generates the fewest delays, on average, workers gain more
using \emph{ATT}.
The reason is that \emph{Forced} minimizes delays, ignoring the costs incurred
by the workers in the process.
To prevent a delay, \emph{Forced} can force a worker to take a large detour and
incur high costs.
It may also take a task from a worker; that worker loses the opportunity to
obtain the reward but is not compensated for the costs that the worker already
incurred.
By contrast, \emph{ATT}  only transfers tasks to workers for whom the promised
reward is higher than the costs for completing the task.

\subsubsection{Individual Profits}

We have seen that both \emph{ATT} and \emph{Forced} result in positive
profits---on average. \linebreak While the mean profit is a useful indicator,
it conveys little information about the fate of individual workers. Even in
scenarios where the mean profit is high, there may exist a notable number of
workers with negative profits.
To gain a deeper insight into the profits of individual workers, we analyze the
profits under \emph{ATT} and \emph{Forced} in more detail.

The cumulative frequency graph in Figure~\ref{fig:worker-profits} shows how the
profits in Scenario~1 under \emph{ATT} and \emph{Forced} are distributed over
the individual participants.
Under the \emph{Forced} strategy, 33.8\% of participants obtained a profit of
EUR 0 or less; i.e.,\ a large fraction of participants would suffer under the
forced task transfers.
In contrast, using \emph{ATT}, only 10.1\% of participants obtained a profit of
EUR 0 ~or less.

Most workers with negative profits under \emph{ATT} are individuals who have
invested costs to perform tasks but were not able to finish them and collect
the reward.
In the absence of task failures (in this experiment, only $0.7\%$ delays
occurred), travel costs are the only source that diminishes a participant's
profit.
Hence, participants with negative profits were not able to fully recover their
travel costs, but \emph{ATT} helped them to avoid penalty payments for failed
tasks.
From the worker perspective, \emph{ATT} is clearly preferable to \emph{Not} and
\emph{Forced}.

\begin{figure}[H]
	\includegraphics[width=1.0\textwidth]{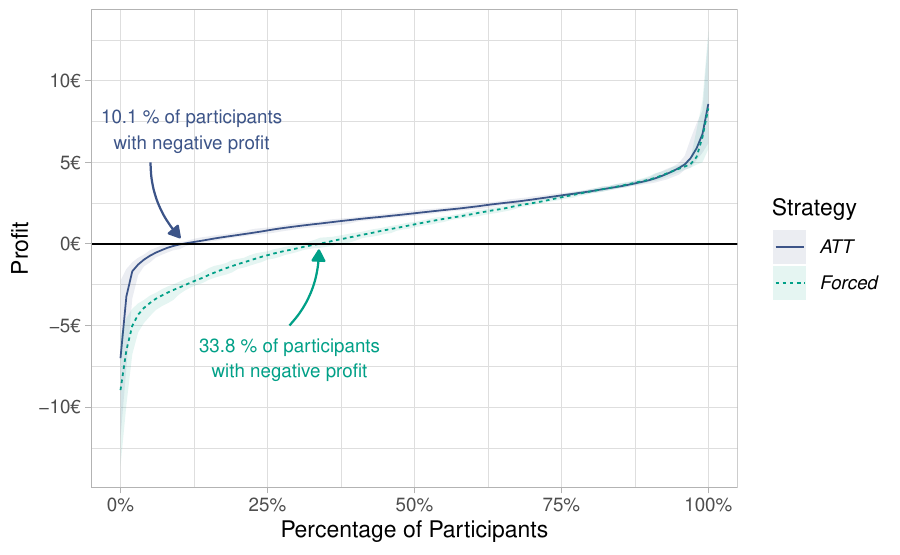}
	\caption[Profit Distribution Under \emph{ATT} and \emph{Forced}
		Strategy in Crowdsensing]{Profit distribution under the \emph{ATT} and
		\emph{Forced} {strategy}  in Scenario~1.}
	\label{fig:worker-profits}
\end{figure}

\section{Conclusions}
\label{sec:conclusion}

Mobile crowdsourcing has proven to be a versatile problem-solving paradigm that
can bring value to many domains where spatial tasks have to be performed
quickly and sustainably at low costs. These tasks are performed by individual
crowdworkers in dynamic and uncertain environments. However, unexpected events,
such as traffic issues or competing obligations, can disrupt tasks  executions,
resulting in unreliable system behavior and low service quality.
In this paper, we argue that supporting peer-to-peer task transfers to
better-suited crowdworkers is a promising approach to prevent task failures. In
particular, we identified two major problems in coordinating crowdworkers in
dynamic environments and discussed corresponding solutions:

\begin{itemize}
	\item	First, crowdworkers must be situation-aware; that is, they must
	      identify emerging issues early enough to trigger task transfers. We proposed a
	      data stream learning approach to predict the probability that a task assigned
	      to a particular crowdworker will be successful. Experimental results showed
	      that various data stream learning methods can be used to predict  service
	      quality with high accuracy.  This enables crowdworkers to make a realistic and
	      continuous assessment of whether a task transfer is useful.

	\item
	      Second, task transfers between crowdworkers must be coordinated.
	      We proposed two  different approaches to task coordination: an opportunistic
	      approach to crowdshipping with collaborative, but non-autonomous workers, and a
	      market-based model with autonomous workers for crowdsensing. Both approaches
	      showed that the service quality could be significantly improved by task
	      transfers.

\end{itemize}

{In} general, through the coordinated interaction of crowdworkers, many task
failures can be prevented, resulting in a more reliable system with higher
stakeholder satisfaction.

From a managerial perspective, the applicability of our approach depends on the
characteristics of the crowd system and the underlying business model. In a
decentralized crowd of individually rational agents, it is important that the
coordination mechanism	respects the autonomy of the workers. Instead, in a
system with top-down management, where agents have agreed to accept orders
(e.g., as employees), the overall quality of the system can be further
optimized. However, some workers may suffer from a decrease in individual
profit. In both cases, we conclude that a good mechanism for predicting task
success is essential.  From a micro-level perspective, crowdworkers would
usually  be interested in maintaining their autonomy. They would prefer the
market-based approach presented, which represents a good compromise between
workers' and system interests.

From a micro-level perspective, crowdworkers would usually be interested in
maintaining their autonomy. They would prefer the market-based approach
presented, which represents a good compromise between worker and system
interests.

In future work, we will focus on some aspects of the practical applicability of
our approach.
Our assumption that task transfers take no time is a simplification of reality
(crowdworkers need to stop, hand over the transfer, etc.).  We will explore how
different notions of transfer costs will affect the task transfer chains. In
this context, we will explore different ways of splitting rewards (penalties)
for (un)successful delivery tasks between the participating crowdworkers  so
that the transfers required by our mechanism are also individually rational for
all stakeholders. This includes investigating the manipulability of information
in general and of the prediction models in particular. We will also  look into
more homogeneous experimental scenarios  to further improve the significance of
the quantitative comparison of the approaches.

Furthermore, the possibility to negotiate handover locations at runtime may
offer advantages. For example, the two workers could meet at a public location
that is close to both of their planned routes. In this case, the detours could
be smaller and handovers faster, resulting in faster and more sustainable
plans.

Our market-based approach uses Vickrey auctions \cite{vickrey1961} for
agreements on a single transfer and a single transfer attribute.  Using a more
sophisticated auction, we could expand our protocol to multidimensional
auctions. Instead of only determining the price by the auction,
multi-dimensional auctions could be used to negotiate further transfer
attributes, such as the transfer deadline and handover location at runtime.
Moreover, combinatorial auctions may bring further benefits. Using
combinatorial auctions, workers could give away and obtain bundles of tasks,
which in some cases can be more attractive than the individual tasks.

\vspace{6pt}

\authorcontributions{Conceptualization, R.B., J.D. (Jeremias {D\"otterl}), J.D.
	({J\"urgen} Dunkel) and S.O.; methodology, R.B., J.D. (Jeremias {D\"otterl}),
	J.D. ({J\"urgen} Dunkel) and S.O.; software, J.D. (Jeremias {D\"otterl});
	validation, R.B., J.D. (Jeremias {D\"otterl}), J.D. ({J\"urgen} Dunkel)  and
	S.O.; formal analysis, R.B., J.D. (Jeremias {D\"otterl}), J.D. ({J\"urgen}
	Dunkel) and S.O.; investigation, R.B., J.D. (Jeremias {D\"otterl}), J.D.
	({J\"urgen} Dunkel), and S.O.; resources, R.B., J.D. (Jeremias {D\"otterl}),
	J.D. ({J\"urgen} Dunkel) and S.O.; data curation, J.D. (Jeremias {D\"otterl});
	writing---original draft preparation, R.B., J.D. (Jeremias {D\"otterl}), J.D.
	({J\"urgen} Dunkel) and S.O.; writing---review and editing, R.B., J.D.
	(Jeremias {D\"otterl}), J.D. ({J\"urgen} Dunkel)  and S.O.; visualization, J.D.
	({J\"urgen} Dunkel); supervision, R.B., J.D. (Jeremias {D\"otterl}) and S.O.
	All authors have read and agreed to the published version of the manuscript.}

\funding{{This} work was supported by the German \emph{Nieders\"achsisches
		Ministerium f\"ur Wissenschaft und Kultur} (MWK) in the programme
	PROFILinternational, as well as the Spanish Ministry of Science and Innovation,
	co-funded by EU FEDER Funds, through grants RTI2018-095390-B-C33,
	PID2021-123673OB-C32 and TED2021-131295B-C33.}

\institutionalreview{Not applicable.}

\informedconsent{Not applicable.}

\dataavailability{The source code used to generate the simulation data study is
	available from the authors upon request.}

\conflictsofinterest{The authors declare no conflict of interest.}

\abbreviations{Abbreviations}{
	The following abbreviations are used in this manuscript:\\

	\noindent
	\begin{tabular}{@{}ll}
		ATT & Auction-based task transfer \\
		FN  & False negative              \\
		FP  & False positive              \\
		HT  & Hoeffding Tree              \\
		KNN & K-nearest neighbors         \\
		RF  & Random Forrest              \\
		TP  & True positive               \\
	\end{tabular}
}

\begin{adjustwidth}{-\extralength}{0cm}

	\reftitle{References}

	\PublishersNote{}
\end{adjustwidth}
\end{document}